\documentclass[conference]{IEEEtran}
\IEEEoverridecommandlockouts
\usepackage{cite}
\usepackage{amsmath,amssymb,amsfonts}
\usepackage{algorithmic}
\usepackage{graphicx}
\usepackage{textcomp}
\usepackage{xcolor}
\usepackage{booktabs}
\usepackage{enumitem}
\usepackage{url}
\usepackage{multirow}
\usepackage{colortbl}
\usepackage{microtype}

\makeatletter
\let\oldthebibliography\thebibliography
\let\endoldthebibliography\endthebibliography
\renewenvironment{thebibliography}[1]{%
  \oldthebibliography{#1}%
  \setlength{\itemsep}{0pt}%
  \setlength{\parskip}{0pt}%
}{\endoldthebibliography}
\makeatother

\begin{document} 

\title{QFoldAgent: An Autonomous Quantum Optimization Multi-Agent System for Protein Structure Prediction}

\author{\IEEEauthorblockN{Winson Chen\IEEEauthorrefmark{1},
Yuqi Zhang\IEEEauthorrefmark{2},
Sixu Chen\IEEEauthorrefmark{2},
Nuo Xu\IEEEauthorrefmark{1},
Qiang Guan\IEEEauthorrefmark{2},
Caiwen Ding\IEEEauthorrefmark{1}}
\IEEEauthorblockA{\IEEEauthorrefmark{1}University of Minnesota, Twin Cities, Minneapolis, United States of America}
\IEEEauthorblockA{\IEEEauthorrefmark{2}Kent State University, Kent, United States of America}
\IEEEauthorblockA{Emails: \{chen9619, xu001536, dingc\}@umn.edu}
\IEEEauthorblockA{Emails: \{yzhan135, schen53, qguan\}@kent.edu}}

\maketitle

\begin{abstract}
Hybrid quantum-classical protein structure prediction depends strongly on Hamiltonian penalty weights, yet existing lattice-based workflows typically fix these coefficients by hand and evaluate only very short fragments in simulation. We present QFoldAgent, a closed-loop multi-agent framework for 5-residue tetrahedral-lattice folding in which a design agent proposes sequence-conditioned penalties, a VQE-based quantum-classical pipeline optimizes the resulting Hamiltonian under Qiskit Aer noise, and a feedback agent uses energy-landscape diagnostics and MolProbity validation signals to refine penalties across cycles. Ground-truth metrics such as RMSD are never exposed to the agents and are used only for evaluation. We study the framework on two complementary datasets: 55 QDockBank-derived fragments with known structures and 100 coverage-optimized unseen sequences. On the QDockBank benchmark, QFoldAgent reduces median RMSD from 3.64 \AA{} to 3.20 \AA{}, with the largest gains on the hardest targets. On unseen sequences, the closed loop raises structural validity from 87.5\% to 98.7\%, recovers 87\% of initially invalid cases, and the strongest controller improves cycle-3 energy on 87\% of sequences while maintaining $>$96\% Ramachandran-favored geometry. These results show that iterative agent control can systematically improve optimization behavior and reduce failure cases in a 5-residue quantum setting.
\end{abstract}


\section{Introduction}

Protein structure prediction remains a central problem in computational biology because molecular structure strongly influences biochemical function, intermolecular binding, and downstream therapeutic design. Recent deep learning systems have substantially advanced the field, with AlphaFold 3~\cite{abramson2024accurate} extending prediction beyond single protein chains to complexes involving proteins, nucleic acids, ligands, ions, and modified residues. Yet important challenges remain, especially in work~\cite{scardino2023good,Perkins-Jechow2025-gd}  involving flexible fragments, ambiguous local geometries, competing low-energy conformations, and structurally constrained functional regions. Recent CASP16~\cite{zhang2025casp16} assessments further indicate that complex structure prediction remains unsolved, with model ranking and stoichiometry prediction continuing to be major bottlenecks. In many cases, these remaining difficulties are not only prediction problems, but also search problems under strong structural constraints.

For quantum computing, these challenging cases can be viewed as constrained optimization problems over discrete structural variables~\cite{robert_resource-efficient_2021}. Rather than treating structure prediction solely as an end-to-end pattern recognition task, selected subproblems can be formulated through a Hamiltonian, which acts as an energy-based objective function that evaluates a candidate conformation according to how well it satisfies encoded structural and physical constraints. This perspective creates a bridge from biological structure prediction to quantum optimization, making selected protein modeling tasks amenable to variational and hybrid quantum-classical solvers. In this setting, the goal is not only to search over candidate conformations, but to construct an objective whose low-energy states align with structurally meaningful solutions.

\begin{figure}
    \centering
    \includegraphics[width=\linewidth]{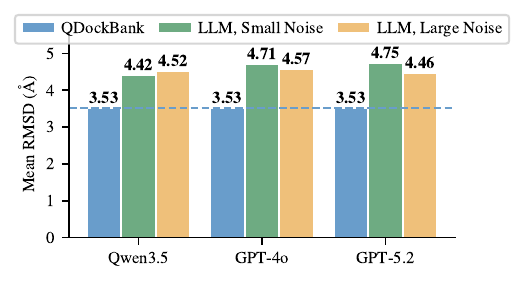}
    \caption{LLM-based mean RMSD vs.\ non-agent QDockBank baseline (dashed line, 3.53~\AA) across 55 proteins. All three LLM produce worse RMSD under both noise conditions, indicating that LLM penalty selection alone cannot outperform the non-agent baseline.}
  \label{fig:llm_vs_qdockbank}
    \label{fig:agent_vs_llm-base}
\end{figure}

In practice, the effectiveness of quantum protein optimization depends critically on how the Hamiltonian penalty weights are chosen. This tuning is typically manual and iterative, creating a major bottleneck in quantum protein structure pipelines. This bottleneck motivates an agentic perspective. Recent work on biological Large Language Model (LLM)-based agents describes a shift from passive predictive models toward systems that can plan, use tools, interpret intermediate outputs, and refine decisions across multi-step scientific workflows \cite{dip2026large,qi2026artificial}. However, current agentic systems are more often used for workflow coordination or algorithmic support than to directly reformulate the mathematical objective of an optimization problem. In this setting, LLM-based approaches remain limited and are unable to effectively solve the task without iterative agentic control. As shown in Figure~\ref{fig:agent_vs_llm-base}, we evaluate three LLMs, Qwen3.5~\cite{qwen3.5}, GPT-4o~\cite{openai2024gpt4ocard}, and GPT-5.2~\cite{openai_gpt52_2025} for Hamiltonian penalty prediction, and benchmark their performance against QDockBank~\cite{zhang2025qdockbank}. Results show that LLM-based approaches underperform QDockBank, exhibiting higher mean RMSD~\cite{Kufareva2012}. This suggests that current LLM-based methods are unable to surpass the manually optimized approaches employed in QDockBank. To address this gap, we introduce \textbf{QFoldAgent}, a closed-loop multi-agent framework to to enable adaptive Hamiltonian refinement for quantum protein structure prediction. Rather than treating Hamiltonian penalty weights as fixed expert-defined choices, QFoldAgent uses a feedback-guided refinement loop: a design agent proposes sequence-conditioned penalties, variational quantum eigensolver (VQE)~\cite{Tilly_2022} optimizer executes the resulting Hamiltonian, and a feedback agent analyzes optimization traces and structural validation signals to refine penalties for the next cycle. The agent serves as a control layer that uses iterative feedback to shape the optimization formulation and adapt the Hamiltonian to each sequence. 

This paper makes three contributions. First, we introduce QFoldAgent, a closed-loop multi-agent framework that treats Hamiltonian penalty tuning in quantum protein structure prediction as an adaptive control problem by linking sequence-conditioned penalty design, VQE execution, structure reconstruction, and feedback-driven refinement across cycles. Second, we establish an evaluation suite setting for this problem using 55 QDockBank-derived fragments with ground-truth structures and 100 coverage-optimized unseen sequences under two Qiskit Aer noise presets. Third, we provide empirical and ablation-based evidence that iterative agent control improves penalty refinement across cycles or independent LLM-based penalty selection, without exposing ground-truth structural metrics to the agent during optimization.

\section{Related Work}

Recent progress in agentic AI suggests that autonomous systems are well suited to scientific workflows requiring iterative reasoning, tool invocation, execution, and verification. Prior studies identify a recurring architectural pattern in which high-level planning is separated from execution and coupled with explicit evaluation or reflection before revision \cite{xu2026agentsurvey}. This principle is also reflected in recent scientific agent frameworks. MCP-SIM adopts a self-correcting loop of planning, acting, reflection, and revision for physics simulation, while VeriMAP combines task decomposition with explicit verification functions for iterative refinement in multi-agent settings \cite{park2026mcpsim,xu2026verimap}. These works position agents as controllers for structured technical pipelines rather than as one-shot text generators.

This perspective is  relevant to quantum computing, where workflows are mathematically structured, tool-dependent, and difficult to automate end to end. Recent studies report growing use of AI and agentic methods across the quantum stack, including software support, calibration, control, and system operation \cite{alexeev2025artificial,sultanow2025quantumagents}. Within this broader trend, LLMs have been explored for tasks such as circuit generation and quantum programming assistance \cite{henderson2025programmingqc}. These systems improve accessibility and support automation, but their role is typically limited to coding, explanation, or localized decision support.

Within quantum protein folding more specifically, prior work has largely focused on encoding strategies and solver design, including quantum annealing for lattice-protein conformations \cite{perdomo2012finding}, resource-efficient gate-model encodings \cite{robert_resource-efficient_2021}, constraint-aware QAOA-style ans\"atze for lattice folding \cite{fingerhuth2018quantum}, and recent peptide-folding studies on quantum computers \cite{boulebnane2023peptide}. This line of work emphasizes encoding choices and solver construction for quantum or hybrid protein-folding formulations, whereas the role of iterative, external control over Hamiltonian parameterization during optimization remains less explored.

However, current LLM support in quantum computing remains constrained. Quantum-Audit shows that quantum computing is a challenging domain for language models because it combines abstract mathematics, non-intuitive physical principles, and rapidly evolving terminology; even strong models exhibit substantial failure modes on advanced expert-authored tasks \cite{afane2026quantumaudit}. More importantly, most existing LLM-based quantum systems remain algorithm-centric: they assist within a predefined workflow rather than adapt the mathematical objective governing the optimization itself \cite{alexeev2025artificial,henderson2025programmingqc}. 

QFoldAgent targets this less explored setting. Consistent with the planner-executor-verifier structure common in agent systems \cite{park2026mcpsim,xu2026verimap}, our framework embeds the agent directly within the optimization loop. Its role is not merely to assist quantum programming or orchestrate a fixed sequence of tools, but to iteratively refine the Hamiltonian used for protein structure prediction based on feedback from the quantum-classical pipeline. The contribution is therefore not simply the use of LLMs in quantum computing, but an agentic mechanism for repeated reformulation of the application objective itself, with the emphasis on workflow adaptation rather than on claiming a new quantum advantage result.

\section{Background}
\label{sec:background}

Protein structure prediction aims to determine the three-dimensional structure adopted by an amino-acid sequence, since structure strongly shapes protein function. Computationally, this problem is difficult because the number of possible conformations grows rapidly with sequence length, while realistic modeling must enforce steric, geometric, and interaction constraints. Current quantum approaches therefore focus on reduced settings such as short fragments, local motifs, binding pockets, or coarse-grained representations, where conformational choices can be expressed in a finite discrete search space while preserving key physical constraints. Candidate conformations are encoded into binary or spin-like variables, and a Hamiltonian is constructed so that low-energy states correspond to plausible structures. The success of the workflow therefore depends heavily on how well the biological problem is mapped to a quantum-optimizable objective.

\begin{figure}[t]
    \centering
    \includegraphics[width=0.8\linewidth]{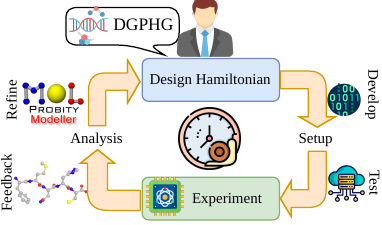}
    \caption{Traditional human-in-the-loop workflow for Hamiltonian refinement. A domain expert manually defines the Hamiltonian and experiment setup, then inspects predicted structures and validation metrics before revising the formulation for the next iteration. Because each round depends on experimental results and expert review, the process is slow and labor intensive.}
    \label{fig:traditional_workflow}
\end{figure}

A conventional quantum workflow for protein structure prediction begins by narrowing the biological task into a quantum-amenable subproblem, since current hardware is better suited to local or coarse-grained formulations than to full all-atom folding at biological scale \cite{robert_resource-efficient_2021,doga2024perspective}. The workflow typically includes selecting a target fragment or motif, defining a discrete structural representation, encoding geometric and physical constraints, constructing a Hamiltonian, mapping it to a quantum-compatible form, running a solver such as VQE or quantum annealing, and decoding and validating the resulting structures. Figure~\ref{fig:traditional_workflow} summarizes this conventional loop. In practice, human experts make most of the key formulation decisions before and after quantum optimization.

Within this process, Hamiltonian design is the central expert task. The Hamiltonian serves as the objective function for the encoded structural problem, so its terms must capture the constraints that distinguish valid from invalid conformations \cite{robert_resource-efficient_2021}. This requires deciding which constraints to include, how to encode them, and how to set penalty coefficients that balance structural realism with tractable optimization. Prior quantum protein studies show that these choices strongly affect the structures a solver can recover \cite{perdomo2012finding,robert_resource-efficient_2021,irback2022annealing,Yan_2023}. In current practice, these decisions are usually made manually through domain knowledge and repeated empirical testing.

This manual formulation step is difficult because small changes in constraint definitions or penalty weights can reshape the energy landscape, affecting feasibility, convergence, and decoded structure quality. Weak penalties may allow invalid conformations, while overly strong penalties can overconstrain the search and hinder optimization~\cite{doga2024perspective}. As a result, the conventional workflow is iterative and expert driven: researchers inspect outputs, revise the Hamiltonian, rerun optimization, and repeat until the formulation behaves acceptably.

In this setting, AI Agent can be introduced as a control layer within the surrounding classical workflow. Recent work shows that AI is increasingly used across quantum computing for preprocessing, control, optimization, and postprocessing, making hybrid classical-AI-quantum pipelines a natural direction for complex technical tasks \cite{alexeev2025artificial}. Our setting builds on this trend by placing AI inside the formulation loop, where it can assist in mapping the biological problem into a quantum objective and refining that objective across iterative cycles.
\section{Methods}

\begin{figure*}[t]
    \centering
    \includegraphics[width=\textwidth]{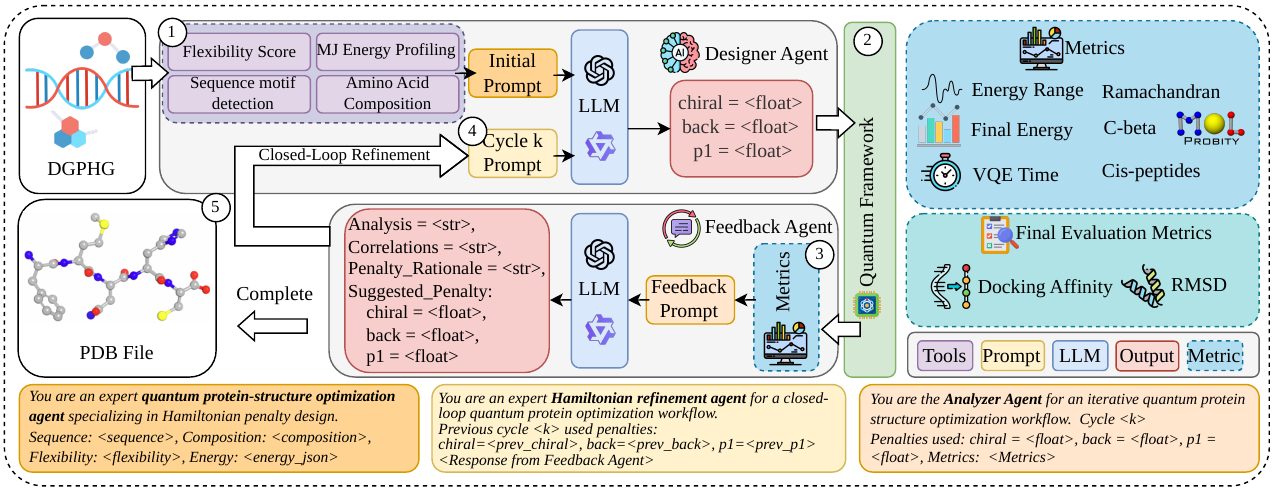}
    \caption{Overview of the closed-loop quantum-classical pipeline for lattice-based protein structure prediction.}
    \label{fig:workflow}
\end{figure*}

We implement a closed-loop pipeline with five stages (Figure~\ref{fig:workflow}). In stage~\textcircled{1}, a \emph{Designer Agent} profiles the input amino acid sequence using composition analysis, motif detection, flexibility scoring, and Miyazawa--Jernigan (MJ) interaction statistics, then proposes initial Hamiltonian penalty weights ($\lambda_{\text{chiral}}$, $\lambda_{\text{back}}$, $\lambda_1$). In stage~\textcircled{2}, the quantum framework compiles the penalized Hamiltonian, performs VQE-based optimization under the configured noise model, decodes the optimized quantum state into lattice conformations via inverse mapping, and reconstructs full-atom candidate structures using MODELLER~\cite{vsali1993comparative}. In stage~\textcircled{3}, Qiskit-derived energy metrics (final energy, convergence quality, energy range) and MolProbity structural validation signals (Ramachandran statistics, C$_\beta$ outliers, twisted peptides) are computed from the candidate structures to summarize current cycle quality. In stage~\textcircled{4}, a \emph{Feedback Agent} receives these optimization and validation metrics together with full cycle history, analyzes how the current penalty settings shaped the energy landscape and structural outcome, and proposes refined penalty values for the next iteration. In stage~\textcircled{5}, this loop repeats for $k$ configured cycles (3 in our experiments), with each cycle refining the Hamiltonian based on the previous cycle's results, before producing the final predicted structure from the best cycle.

\subsection*{Problem Mapping and Quantum Encoding}

Following QDockBank, we formulate fragment-level protein structure prediction as sequence-conditioned optimization on a tetrahedral lattice. Each residue is represented as a node with four allowed extension directions, fixed bond length, and an approximately $109.4^\circ$ bond angle, preserving key stereochemical properties of the backbone \cite{Yan_2023}. This encoding maps each candidate conformation to a quantum state, with conformation energy represented by the expectation of a compiled Hamiltonian. In QDockBank notation, the total Hamiltonian is defined as
\begin{equation}
H_t=\lambda_c H_c+\lambda_g H_g+\lambda_d H_d+\lambda_i H_i,
\end{equation}
where $H_c$ enforces chirality constraints, $H_g$ enforces backbone geometry constraints, $H_d$ penalizes steric conflicts, and $H_i$ captures the sequence-dependent Miyazawa--Jernigan (MJ) interaction term. The coefficients $\lambda_c, \lambda_g, \lambda_d,$ and $\lambda_i$ are tunable scalar penalty weights that balance the relative contribution of chirality, backbone geometry, collision avoidance, and residue interaction terms in the total Hamiltonian. Our implementation keeps this decomposition but exposes only three adaptive coefficients to the agent:
\[
\lambda_c \mapsto \lambda_{\text{chiral}},\quad
\lambda_g \mapsto \lambda_{\text{back}},\quad
\lambda_d \mapsto \lambda_1,\quad
\lambda_i = 1 \; \text{(fixed)}.
\]
Thus the agent changes only the relative strength of the three constraint penalties; the MJ interaction term is compiled from the sequence and remains fixed during all feedback cycles.

\subsection*{Stage \textcircled{1}: Design Agent}

At the start of each cycle, the Hamiltonian design agent selects penalty weights that balance interaction terms and hard constraints. The initialization stage characterizes the input sequence using four profiling modules:
\begin{enumerate}[label=(\alph*)]
  \item \textbf{Amino acid composition analysis:} residues are categorized as hydrophobic ($F, I, L, V, W, M$), polar ($S, T, N, Q, C, Y$), or charged ($D, E, K, R, H$). The profiler also flags proline, cysteine, and glycine content.
  \item \textbf{Sequence motif detection:} detects hydrophobic runs (length $\geq 3$), cysteine pairs, and helix-breaking proline positions.
  \item \textbf{Flexibility scoring:} sequence flexibility is estimated on a normalized $[0,1]$ scale from glycine/proline content.
  $$\text{flexibility} = 0.5 + 0.15 \times \frac{N_{\text{Gly}}}{L} - 0.2 \times \frac{N_{\text{Pro}}}{L}$$
  \item \textbf{Miyazawa--Jernigan (MJ) energy profiling:} pairwise contact tendencies are estimated from MJ statistics, including total and mean interaction energy and the count of strongly favorable pairs.
  $$E_{\text{total}} = \sum_{i<j} \text{MJ}(a_i, a_j)$$
\end{enumerate}

These features are passed to the LLM together with qubit-budget metadata (initial qubits $=4(N-1)^2$, then reduced by unused-qubit elimination). The LLM proposes three penalties:
\begin{itemize}
  \item \texttt{penalty\_chiral} for chirality constraints,
  \item \texttt{penalty\_back} for consecutive same-axis turns,
  \item \texttt{penalty\_1} for local overlap constraints.
\end{itemize}

Using the penalties, we construct the total Hamiltonian as
\begin{equation}
\begin{aligned}
H_t=\;&H_{\text{MJ}}+\lambda_{\text{chiral}} H_{\text{chiral}}
+\lambda_{\text{back}} H_{\text{back}}+\lambda_1 H_{\text{overlap}},
\end{aligned}
\end{equation}
where $H_{\text{overlap}}$ aggregates the local steric-conflict terms used in the tetrahedral lattice construction. The solver does not operate on a simple ``two qubits per turn label'' register. Instead, the tetrahedral encoding expands the turn decisions into a binary indicator register whose pre-compression size scales as $4(N-1)^2$. For the 5-residue fragments used throughout this paper, this corresponds to 64 qubits before eliminating fixed or unused variables. After compression, the working optimization register is approximately 22 qubits, and the backend requirement is approximately 27 qubits once the 5-qubit execution buffer used by our VQE stack is included.

Here,
\begin{equation}
H_{\text{back}} = \sum_i \sum_{k=0}^{3} \hat{\delta}_k^{(i)}\,\hat{\delta}_k^{(i+1)},
\end{equation}
where $\hat{\delta}_k^{(i)}$ is an indicator projector for turn direction $k$ at position $i$. Identity-only or symmetry-fixed variables are removed after construction, so the optimized circuit acts only on the compressed register rather than the full indicator expansion.

For cycle $k>1$, the LLM receives previous-cycle diagnostics from the Feedback Agent (Section~\ref{sec:feedback-analysis}) (energy range, convergence quality, and structural-validation summary) and updates penalties without recomputing sequence features.

\subsection*{Stages \textcircled{2}--\textcircled{3}: Quantum Folding Pipeline}

\begin{figure}[t]
    \centering
    \includegraphics[width=0.4\textwidth]{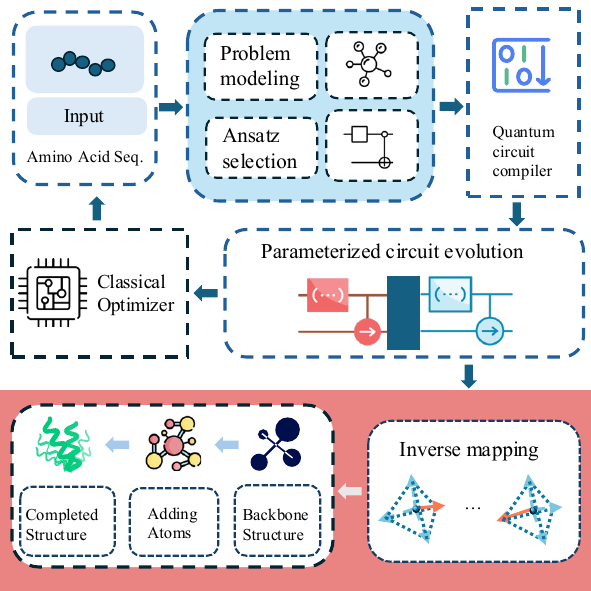}
    \caption{Quantum folding pipeline. The amino acid sequence is encoded into a tetrahedral-lattice Hamiltonian, optimized with VQE using a parameterized quantum circuit and COBYLA, then decoded into backbone conformations and reconstructed into full-atom structures with MODELLER.}
    \label{fig:quantum_workflow}
\end{figure}
\subsubsection{Variational Optimization with VQE}

As shown in Figure~\ref{fig:quantum_workflow}, the compiled Hamiltonian $H_t$, we solve the corresponding ground-state search problem using a Variational Quantum Eigensolver (VQE), consistent with QDockBank. Starting from an initial state $\lvert \psi_i \rangle$, the parameterized circuit $U(\boldsymbol{\theta})$ prepares
\begin{equation}
\lvert \psi(\boldsymbol{\theta}) \rangle = U(\boldsymbol{\theta}) \lvert \psi_i \rangle,
\end{equation}
with variational objective
\begin{equation}
E(\boldsymbol{\theta})=\langle \psi_i \vert U^\dagger(\boldsymbol{\theta}) H_t U(\boldsymbol{\theta}) \vert \psi_i \rangle.
\end{equation}
Minimizing $E(\boldsymbol{\theta})$ searches for low-energy states whose decoded bit strings correspond to lattice conformations satisfying the encoded folding constraints.

We use an EfficientSU2 ans\"atz with alternating single-qubit rotations ($R_Y$, $R_Z$) and CNOT entanglers on $n_{\text{compressed}}+5$ qubits. In the 5-residue setting, the compiled circuit has approximately 18 layers after decomposition and 56 trainable parameters. Parameters $\boldsymbol{\theta}$ are initialized randomly and optimized with COBYLA \cite{powell1994direct} for up to 50 iterations, using 200 shots per energy evaluation.

This VQE budget is held fixed across both experiments to isolate the effect of Hamiltonian refinement from differences in quantum optimization effort. The chosen budget should therefore be interpreted as a controlled evaluation setting rather than a claim of full VQE convergence. To reduce sensitivity to local minima, we retain all intermediate $(E_i,\boldsymbol{\theta}_i)$ pairs and keep the six lowest-energy parameter sets for downstream decoding. The pipeline supports both IBM Quantum backends, using least-busy selection and level-3 transpilation, and Qiskit Aer simulation with a matrix-product-state backend and maximum bond dimension 512.

\subsubsection{Measurement and Structural Interpretation}

After optimization, each retained low-energy solution is measured with 100{,}000 shots to estimate
\begin{equation}
P(\mathbf{b}) = \left|\langle \mathbf{b} \mid \psi(\boldsymbol{\theta}^\ast) \rangle\right|^2.
\end{equation}
This larger shot budget is used only for final decoding, allowing a more reliable estimate of the output bit-string distribution than in the lower-shot optimization loop.

The most probable bit string $\mathbf{b}^\ast$ is decoded into turn variables $Q_i \in \{0,1,2,3\}$ and mapped to C$_\alpha$ coordinates as
\begin{equation}
\mathbf{r}_{i+1} = \mathbf{r}_i + (-1)^i\,\mathbf{v}_{Q_i}.
\end{equation}
The resulting coordinates are rescaled to the canonical C$_\alpha$--C$_\alpha$ distance of 3.8~\AA{} to produce a coarse backbone trace.

\subsubsection{Atomic Reconstruction and Structural Validation}

Predicted C$_\alpha$ traces are reconstructed into all-atom models with standard amino acid templates using MODELLER. Backbone geometry is normalized to standard bond lengths and angles so that the resulting structures can be evaluated for stereochemical plausibility. To assess functional compatibility after atomic reconstruction, we additionally use AutoDock Vina~\cite{autodock} to estimate the ligand docking affinity of each candidate structure. We then compute structural validation signals including Ramachandran favored percentage, C$_\beta$ outliers, and twisted peptides using MolProbity~\cite{molprobity}. In Experiment~1, RMSD to the ground-truth fragment is used as the primary endpoint metric. Because low variational energy does not always yield the best decoded structure, the pipeline evaluates candidates using both optimization and downstream structural criteria. These structural signals are subsequently passed to the feedback stage for the next cycle of Hamiltonian refinement.

\subsection*{Stage \textcircled{4}: Feedback Agent}
\label{sec:feedback-analysis}

This agent uses an \textbf{LLM-as-judge}~\cite{zheng2023judgingllmasajudgemtbenchchatbot} pattern to evaluate VQE output metrics and decide how to refine penalty parameters. The LLM receives the configured metric set and acts as the primary decision-maker for penalty adjustments. The metrics presented to the LLM judge are organized into modular categories:

\begin{itemize}
  \item \textbf{Energy landscape:} convergence quality score $q \in [0,1]$, computed as $q = 0.4\,q_{\text{imp}} + 0.35\,q_{\text{stab}} + 0.25\,q_{\text{mono}}$, where $q_{\text{imp}} = (E_{\text{init}} - E_{\text{final}})/|E_{\text{init}}|$ measures improvement ratio, $q_{\text{stab}} = 1 - \min(\sigma_{\text{tail}}/\mu_{\text{tail}}, 1)$ measures late-iteration stability over the final 20\% of iterations, and $q_{\text{mono}}$ is the fraction of iterations with decreasing energy. The agent also receives initial/final/min/max VQE energies and the energy range ($E_{\max} - E_{\min}$).
  \item \textbf{Structural validation:} Ramachandran favored/outlier statistics, C$_\beta$ deviation counts, and twisted-peptide counts from the reconstructed full-atom model. These metrics validate \emph{local} backbone geometry rather than global fold accuracy.
  \item \textbf{Structural summary:} 3D C$_\alpha$ coordinates, pairwise distance matrix, and geometry statistics derived from the reconstructed structure.
\end{itemize}

The LLM judge produces a structured output: (1)~\texttt{[ANALYSIS]}---assessment of current structural quality; (2)~\texttt{[CORRELATION]}---how penalties shaped the energy landscape; (3)~\texttt{[PENALTY\_RATIONALE]}---specific reasoning for each suggested adjustment; and (4)~\texttt{[PENALTIES]}---refined values for $(\lambda_{\text{chiral}}, \lambda_{\text{back}}, \lambda_1)$. The prompt includes the physical meaning of each penalty, full cycle history, and metric interpretation guides. Guidance adapts by cycle number: bold adjustments early (exploration), moderate in middle cycles, and small targeted adjustments late (fine-tuning).

\subsection*{Stage \textcircled{5}: Final Structure Selection}

After $k$ cycles are completed, the pipeline selects the final predicted structure. For each cycle, the VQE retains multiple low-energy candidate conformations (Section~III-C). The pipeline compares all candidates across all cycles and selects the structure from the best-performing cycle based on the lowest VQE final energy. This structure is output as the final PDB file, which includes the full-atom reconstruction produced by MODELLER from the selected cycle's C$_\alpha$ trace. Evaluation-only metrics (RMSD in Experiment~1) are computed on this final output but are never fed back to the agents.

\section{Experiments}
\label{sec:experiments}

To evaluate how effectively the closed-loop pipeline adapts Hamiltonian penalties across sequences and noise conditions, we test it on two complementary settings. The 55 QDockBank protein fragments with known ground-truth structures allow direct measurement of structural accuracy; 100 newly generated unseen 5-residue sequences probe whether the agent's penalty-refinement strategy generalizes when no reference structure is available. Together with the evaluation scripts and baseline results released alongside this work, these two tracks form a reusable evaluation suite for benchmarking future LLM controllers in quantum optimization workflows.

\subsection{Experimental Setup}
\label{sec:setup}

\textit{Models and infrastructure.}
We evaluate three LLM controllers as the reasoning backbone for both agents: \textbf{OS} (Qwen3.5-35B-A3B~\cite{qwen3.5}, served locally via vLLM), \textbf{SP} (GPT-4o~\cite{openai2024gpt4ocard}, OpenAI API), and \textbf{SP+} (GPT-5.2~\cite{openai_gpt52_2025}, OpenAI API). The Hamiltonian Designer temperature is set to $T{=}0.6$ (encouraging exploration) and the Feedback Analyzer to $T{=}0.3$ (deterministic assessments). All runs use an EfficientSU2 ans\"atz with COBYLA optimizer, 50 iterations, and 200 measurement shots per energy evaluation. This fixed compute budget is shared across both experiments so that outcome differences reflect LLM-driven penalty adaptation rather than variation in optimization effort. All experiments run on a single server with dual AMD EPYC 9354 processors (128 threads), 756~GB RAM, and 8$\times$ NVIDIA RTX PRO 6000 Blackwell (96~GB each), running Ubuntu 22.04. Quantum simulation uses Qiskit Aer on CPU only (\texttt{CUDA\_VISIBLE\_DEVICES=""}) to ensure reproducibility, while the GPUs are reserved exclusively for local vLLM inference of the Qwen3.5 controller.  

\textit{Quantum noise simulation.}
Both experiments are evaluated under two Qiskit Aer noise presets---\textit{Small} and \textit{Large}---designed to bracket the error regimes of current superconducting quantum processors. Both include thermal relaxation with gate times of 50~ns (single-qubit) and 300~ns (two-qubit). The \textit{Small} preset uses depolarizing rates of $2 \times 10^{-4}$ (1Q) and $2 \times 10^{-3}$ (2Q), readout error $1 \times 10^{-2}$, and $T_1/T_2 = 50/70~\mu\text{s}$, representing a gate-error-dominated regime. The \textit{Large} preset increases all noise channels to $5 \times 10^{-4}$ (1Q) and $5 \times 10^{-3}$ (2Q), readout error $1.5 \times 10^{-2}$, and $T_1/T_2 = 80/60~\mu\text{s}$, where the shorter $T_2$ shifts the dominant error source toward decoherence.

\textit{Evaluation metrics.}
We use three categories of metrics:

\begin{itemize}[leftmargin=1.2em,itemsep=2pt,topsep=2pt]
  \item \textbf{Endpoint accuracy}: RMSD against ground-truth C$_\alpha$ positions (QDockBank set); MolProbity-based structural validity---Ramachandran favored ${\geq}$95\%, zero C$_\beta$ outliers, zero twisted peptides (unseen-sequence set).
  \item \textbf{Optimization process}: the normalized energy-to-penalty ratio $\text{E/P} = E_{\mathrm{final}}/\Sigma\lambda$ isolates optimization quality from penalty scale; a monotonic decrease across cycles indicates directed refinement. Also reported: energy range, VQE convergence rate, and penalty trajectory.
  \item \textbf{Structural quality} (secondary): Ramachandran favored \%, convergence quality, and docking affinity.
\end{itemize}

All paired comparisons use the Wilcoxon signed-rank test (two-sided), with rank-biserial correlation $r$ as effect size ($|r| > 0.5$: large effect)~\cite{wilcoxon}.

\subsection{QDockBank Benchmark}
\label{sec:exp1}

\begin{figure}[t]
    \centering
    \includegraphics[width=\columnwidth]{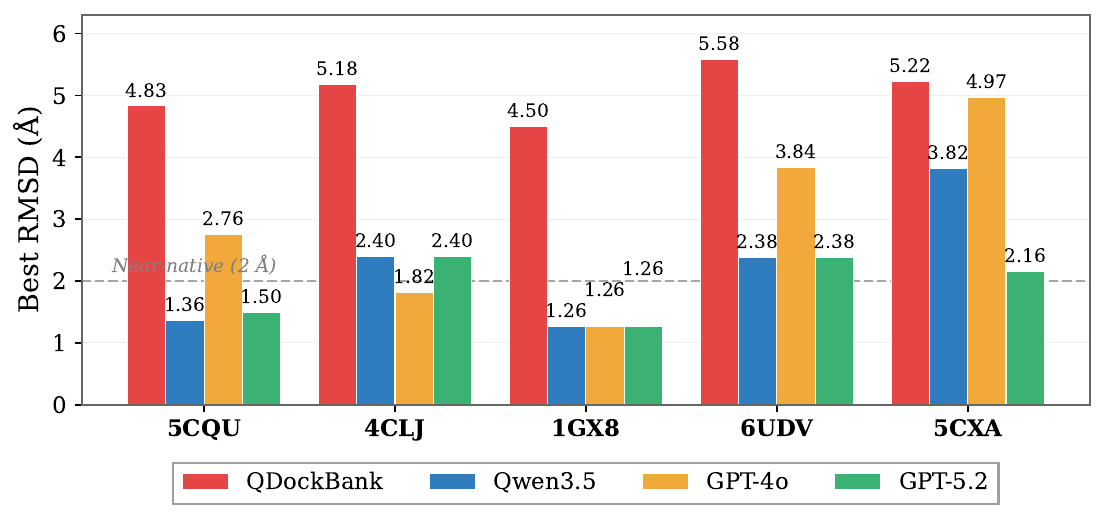}
    \caption{Best RMSD (\AA) on the five QDockBank targets with the largest improvement under small Aer noise. The agent drives several difficult cases below 2~\AA{} where the fixed-parameter baseline exceeds 4~\AA{}.}
    \label{fig:rmsd_bar}
\end{figure}

\textit{Setup.}
We evaluate QFoldAgent on 55 protein--ligand complexes from QDockBank using 5-residue binding-site fragments, comparing the three LLM controllers against the \textbf{QDockBank baseline}, which uses fixed penalty weights and a single optimization pass. QDockBank reports RMSD on variable-length fragments (5--14 residues), whereas our agent operates on fixed 5-residue sub-fragments; the comparison is therefore informative for workflow-level improvement but is not a matched-length benchmark. Each model runs \textbf{3 closed-loop cycles} per target under both Aer noise presets. The primary metric is RMSD after Kabsch alignment~\cite{kabsch1976solution}; values below 2~\AA{} are generally considered near-native for short fragments. When the predicted fragment is shorter than the ground-truth PDB, RMSD is computed on the matching C$_\alpha$ subset.

\subsubsection{Results on the QDockBank baseline}

QFoldAgent improves on the fixed-parameter QDockBank baseline on many targets under both noise settings, with the strongest gains concentrated on the hardest cases. The QDockBank reference has median RMSD \textbf{3.64~\AA} and mean RMSD \textbf{3.53~\AA}. Per-model win counts are \textbf{27/55} (Qwen3.5), \textbf{29/55} (GPT-4o), and \textbf{33/55} (GPT-5.2) under small noise, and \textbf{32/55} (Qwen3.5), \textbf{28/55} (GPT-4o), and \textbf{28/55} (GPT-5.2) under large noise. As a post-hoc upper bound (not deployable), an oracle selecting the best model per target improves on \textbf{40/55} targets under small noise and \textbf{39/55} under large noise; the failures are concentrated on already-easy targets where the baseline is already below 3.0~\AA. The oracle result suggests that combining multiple models could yield further gains, since different controllers succeed on different targets. For any single model, however, each individually beats the fixed baseline on roughly half or more of targets, and all three land in a narrow RMSD band (median 3.20--3.33~\AA), suggesting that the benefit is not tied to any particular backend alone. The stronger evidence for a framework-level effect comes from the behavioral comparison against the LLM-only baseline presented below.


 The advantage is most visible on the hardest QDockBank cases. Under small noise, Figure~\ref{fig:rmsd_bar} highlights the five largest improvements: 5CQU                 
  ($4.83{\to}1.36$~\AA), 4CLJ ($5.18{\to}1.82$~\AA), 1GX8 ($4.50{\to}1.26$~\AA), 6UDV ($5.58{\to}2.38$~\AA), and 5CXA ($5.22{\to}2.16$~\AA). These are precisely the cases  
  where iterative penalty adaptation is expected to help most, since the fixed-parameter baseline fails to locate a near-native fold. Large-noise results show the same     
  trend, with strong recoveries on 1GX8 ($4.50{\to}0.62$~\AA) and 6G98 ($3.69{\to}0.85$~\AA). The strongest individual cases reach sub-1~\AA{} RMSD under both noise        
  presets, including 3EAX ($2.06{\to}0.43$~\AA) and 4ZB8 ($3.36{\to}0.89$~\AA).    

\subsubsection{Directed optimization evidence}

\begin{figure}[t]
    \centering
    \includegraphics[width=\columnwidth]{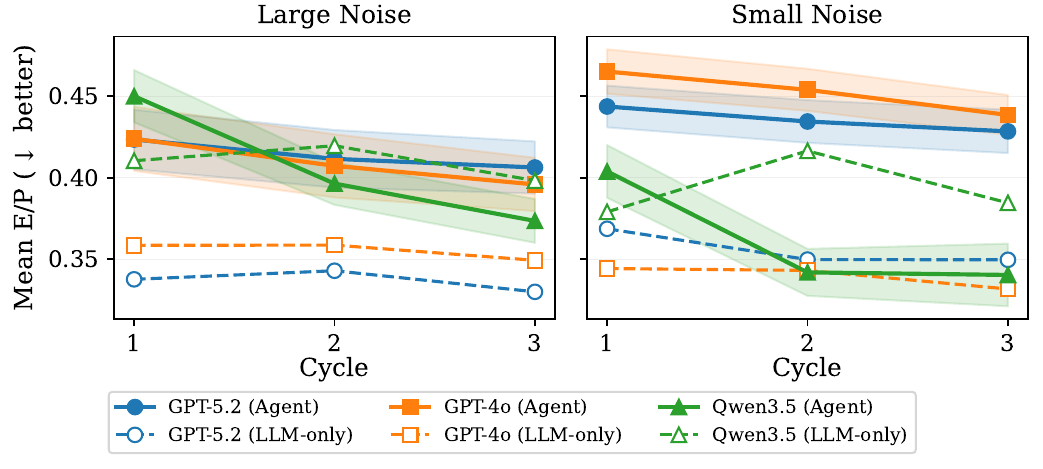}
    \caption{E/P trajectory on the 55 ground-truth targets. Agent configurations (solid) improve systematically from cycle~1 to cycle~3; LLM-only baselines (dashed) show no directed trend.}
    \label{fig:ep_trajectory_55gt}
\end{figure}

To test whether the improvement reflects directed optimization rather than random sampling, we compare behavioral trajectories between agent and LLM-only pipelines on the same 55 targets, where both run 3 independent cycles. The normalized energy-efficiency ratio $\text{E/P} = E_{\mathrm{final}} / \Sigma\lambda$ isolates optimization quality from penalty scale; directed optimization should produce monotonic E/P improvement across cycles, whereas random sampling should show no systematic trajectory. E/P improves from cycle~1 to cycle~3 in \textbf{5/6} agent model--noise configurations (Wilcoxon $p < 0.05$, rank-biserial $r = 0.40$--$0.86$), while LLM-only independent sampling shows no directed trend (\textbf{0/6} significant, all $p > 0.28$). Figure~\ref{fig:ep_trajectory_55gt} visualizes these contrasting trajectories. Penalty convergence provides complementary evidence: the ratio $\max(\lambda)/\min(\lambda)$ decreases in \textbf{5/6} agent configurations ($p < 0.05$) vs.\ \textbf{0/6} for LLM-only. The same LLM, given the same number of cycles without structured feedback, does not develop a directed penalty strategy---isolating the feedback loop as the causal mechanism.

\begin{table}[t]
\caption{Structural quality metrics on the QDockBank set, averaged over 55 targets (best-RMSD cycle per protein). Rama = Ramachandran favored; Conv.\ Q = convergence quality; Dock = AutoDock Vina affinity in kcal/mol (more negative = stronger binding).}
\label{tab:exp1_quality}
\centering
\setlength{\tabcolsep}{3pt}
\begin{tabular}{@{}llcccc@{}}
\toprule
Model & Noise & Rama (\%) & Conv.\ Q & Dock & Med.\ RMSD \\
\midrule
Qwen3.5 & small & 97.0 & 0.650 & $-$3.67 & 3.33 \\
GPT-4o  & small & 97.0 & 0.651 & $-$3.68 & 3.28 \\
GPT-5.2 & small & 97.6 & 0.654 & $-$3.69 & 3.20 \\
\midrule
Qwen3.5 & large & 97.6 & 0.660 & $-$3.74 & 3.20 \\
GPT-4o  & large & 99.4 & 0.651 & $-$3.71 & 3.28 \\
GPT-5.2 & large & 97.0 & 0.653 & $-$3.75 & 3.20 \\
\midrule
\multicolumn{2}{l}{QDockBank ref.} & --- & --- & --- & 3.64 \\
\bottomrule
\end{tabular}
\end{table}

\subsubsection{Structural quality and endpoint ceiling}

The reduction in RMSD does not come at the expense of backbone geometry. Table~\ref{tab:exp1_quality} summarizes Ramachandran favored percentages, energy convergence quality, and docking affinity across all six configurations. Ramachandran percentages remain above 97\% in every setting, convergence quality is tightly clustered around 0.65, and docking affinity is stable between $-3.67$ and $-3.75$~kcal/mol, confirming that the RMSD improvements do not compromise binding-pocket geometry.

\begin{table}[t]
\caption{Per-protein best RMSD (\AA) under small noise: QDockBank baseline vs.\ agent pipeline (3 cycles). \underline{\textbf{Bold}} = best per protein; agent improves on 40/55.}
\label{tab:small_noise_vs_qdockbank}
\centering
\scriptsize
\setlength{\tabcolsep}{2pt}
\renewcommand{\arraystretch}{0.94}
\begin{tabular}{llcccc@{\hspace{4pt}}|@{\hspace{4pt}}llcccc}
\toprule
PDB & Seq & QDB & OS & SP & SP+ & PDB & Seq & QDB & OS & SP & SP+ \\
\midrule
1e2k & DGPHG & 2.32 & 4.53 & 4.53 & \underline{\textbf{2.28}} & 4cig & VRDQA & 3.81 & 2.87 & \underline{\textbf{2.73}} & 2.87 \\
1e2l & AQITM & 3.54 & \underline{\textbf{3.42}} & 4.07 & 4.05 & 4clj & ILMEL & 5.18 & 2.40 & \underline{\textbf{1.82}} & 2.40 \\
1gx8 & SAPLR & 4.50 & \underline{\textbf{1.26}} & \underline{\textbf{1.26}} & \underline{\textbf{1.26}} & 4f5y & GLAWS & 3.20 & \underline{\textbf{1.26}} & \underline{\textbf{1.26}} & 3.80 \\
1hdq & SIHSY & \underline{\textbf{2.85}} & 3.69 & 3.33 & 3.33 & 4fp1 & PVHTA & 5.04 & \underline{\textbf{2.91}} & 3.21 & \underline{\textbf{2.91}} \\
1m7y & TAGAT & 3.19 & \underline{\textbf{2.44}} & 3.28 & 3.14 & 4jpx & DYLEA & 4.35 & 5.80 & \underline{\textbf{3.96}} & \underline{\textbf{3.96}} \\
1ppi & PWWER & \underline{\textbf{2.68}} & 3.05 & 3.53 & 3.53 & 4jpy & YLEAY & 3.85 & 4.26 & \underline{\textbf{1.87}} & 2.35 \\
1qin & QQTML & \underline{\textbf{2.75}} & 2.90 & 2.90 & 4.79 & 4mc1 & DTGAD & 3.90 & \underline{\textbf{2.49}} & 4.73 & 2.50 \\
1yc4 & ELISN & 4.09 & \underline{\textbf{3.72}} & \underline{\textbf{3.72}} & 4.69 & 4mo4 & NIGGF & 2.00 & 2.48 & 2.48 & \underline{\textbf{1.91}} \\
1zsf & LLDTG & \underline{\textbf{2.21}} & 2.87 & 5.00 & 4.08 & 4q87 & SLTTP & 3.64 & \underline{\textbf{1.89}} & \underline{\textbf{1.89}} & \underline{\textbf{1.89}} \\
2avo & LIDTG & \underline{\textbf{3.37}} & 4.54 & 3.98 & 6.42 & 4tmk & IEGLE & 4.35 & \underline{\textbf{1.33}} & \underline{\textbf{1.33}} & 3.19 \\
2bfq & AFPAV & 4.46 & \underline{\textbf{1.96}} & 3.43 & 2.31 & 4xaq & GSYSD & 3.24 & \underline{\textbf{1.67}} & 2.15 & \underline{\textbf{1.67}} \\
2bok & EDACQ & 3.33 & 3.50 & \underline{\textbf{1.92}} & 2.71 & 4y79 & ACQGD & 2.81 & \underline{\textbf{2.18}} & 4.21 & 3.28 \\
2qbs & HCSAG & \underline{\textbf{2.43}} & 4.07 & 5.24 & 3.39 & 4zb8 & GGPNG & 3.36 & \underline{\textbf{0.89}} & 2.44 & 2.44 \\
2v25 & ATFTI & \underline{\textbf{1.66}} & 7.51 & 2.64 & 4.03 & 5c28 & CDLCS & 3.53 & 3.91 & 4.53 & \underline{\textbf{1.43}} \\
2vwo & DACQG & \underline{\textbf{3.39}} & 5.75 & 3.40 & 5.91 & 5cqu & RKLGR & 4.83 & \underline{\textbf{1.36}} & 2.76 & 1.50 \\
2xxx & GAVED & 3.75 & 3.98 & \underline{\textbf{1.51}} & 3.45 & 5cxa & FDGKG & 5.22 & 3.82 & 4.97 & \underline{\textbf{2.16}} \\
3b26 & LISNS & 3.63 & \underline{\textbf{1.48}} & 3.61 & 3.61 & 5kqx & LLNTG & 3.77 & 4.37 & \underline{\textbf{2.48}} & 2.80 \\
3ckz & VKDRS & \underline{\textbf{0.83}} & 3.00 & 6.01 & 4.14 & 5kr2 & LNTGA & 4.95 & 4.22 & 4.22 & \underline{\textbf{4.07}} \\
3d7z & YLVTH & 5.48 & 3.74 & \underline{\textbf{2.78}} & 3.79 & 5nkb & MIITE & 5.43 & 6.89 & 8.14 & \underline{\textbf{5.06}} \\
3d83 & LVTHL & 3.75 & 2.43 & \underline{\textbf{2.34}} & \underline{\textbf{2.34}} & 5nkc & IITEY & 4.28 & \underline{\textbf{1.85}} & 1.99 & 1.99 \\
3dx3 & HNDPG & 2.30 & 3.20 & \underline{\textbf{2.03}} & \underline{\textbf{2.03}} & 5nkd & ITEYM & 4.10 & \underline{\textbf{3.14}} & \underline{\textbf{3.14}} & \underline{\textbf{3.14}} \\
3eax & RYRDV & 2.06 & 3.75 & 2.83 & \underline{\textbf{0.43}} & 5tya & LTTPP & 3.67 & \underline{\textbf{2.39}} & 4.16 & 4.46 \\
3ibi & IQFHF & \underline{\textbf{1.91}} & 6.44 & 4.25 & 6.44 & 6czf & LRKAN & \underline{\textbf{2.91}} & 5.93 & 3.69 & 3.20 \\
3nxq & VCHAS & \underline{\textbf{2.97}} & 5.02 & 5.13 & 5.13 & 6ezq & AKQRL & 3.86 & 4.10 & \underline{\textbf{2.26}} & 5.56 \\
3s0b & GIKAV & \underline{\textbf{2.89}} & 4.81 & 5.93 & 4.81 & 6g98 & RNNGH & \underline{\textbf{3.69}} & 7.20 & 7.20 & 4.58 \\
3tcg & IEGVP & \underline{\textbf{1.71}} & 1.91 & 1.77 & 1.91 & 6p86 & VYSSG & 4.10 & \underline{\textbf{3.30}} & 7.55 & 5.34 \\
3vf7 & LDTGA & 3.79 & 3.85 & \underline{\textbf{2.71}} & \underline{\textbf{2.71}} & 6udv & SLSRV & 5.58 & \underline{\textbf{2.38}} & 3.84 & \underline{\textbf{2.38}} \\
4aoi & VVLPY & 3.93 & \underline{\textbf{1.84}} & \underline{\textbf{1.84}} & 3.56 & \multicolumn{2}{>{\cellcolor{gray!20}}l}{\textit{RMSD $<$ QDB}} & \cellcolor{gray!20}-- & \cellcolor{gray!20}27 & \cellcolor{gray!20}29 & \cellcolor{gray!20}33 \\
\bottomrule
\end{tabular}
\end{table}

Per-protein Wilcoxon tests against the QDockBank reference do not reach significance for any model--noise configuration ($p > 0.05$). Across all 13 pipeline configurations, mean best RMSD falls within a \textbf{0.29~\AA} band (3.24--3.53~\AA), indicating a 5-residue structural ceiling at which method discrimination collapses. The QDockBank evaluation itself reports that AlphaFold3 achieves better RMSD than the quantum baseline on \textbf{90\%} of S-group fragments (5--8 residues) but only \textbf{58.7\%} of L-group fragments (13--14 residues), confirming that method discrimination weakens with decreasing fragment length. At this scale, the agent's contribution is therefore better characterized as improved optimization behavior and reduced failure cases without compromising structural quality, rather than as a statistically significant endpoint advantage.

\begin{table}[t]
\caption{Per-protein best RMSD (\AA) under large noise: QDockBank baseline vs.\ agent pipeline (3 cycles). \underline{\textbf{Bold}} = best per protein; agent improves on 39/55.}
\label{tab:large_noise_vs_qdockbank}
\centering
\scriptsize
\setlength{\tabcolsep}{2pt}
\renewcommand{\arraystretch}{0.94}
\begin{tabular}{llcccc@{\hspace{4pt}}|@{\hspace{4pt}}llcccc}
\toprule
PDB & Seq & QDB & OS & SP & SP+ & PDB & Seq & QDB & OS & SP & SP+ \\
\midrule
1e2k & DGPHG & 2.32 & \underline{\textbf{2.19}} & 4.53 & 3.29 & 4cig & VRDQA & 3.81 & 2.87 & 3.83 & \underline{\textbf{2.52}} \\
1e2l & AQITM & \underline{\textbf{3.54}} & 4.07 & 4.64 & 4.64 & 4clj & ILMEL & 5.18 & 2.40 & \underline{\textbf{1.82}} & \underline{\textbf{1.82}} \\
1gx8 & SAPLR & 4.50 & 1.26 & \underline{\textbf{0.62}} & 1.76 & 4f5y & GLAWS & 3.20 & 3.74 & \underline{\textbf{1.26}} & \underline{\textbf{1.26}} \\
1hdq & SIHSY & 2.85 & 3.33 & \underline{\textbf{2.16}} & 3.33 & 4fp1 & PVHTA & 5.04 & 3.37 & \underline{\textbf{2.91}} & 3.21 \\
1m7y & TAGAT & 3.19 & \underline{\textbf{2.29}} & \underline{\textbf{2.29}} & \underline{\textbf{2.29}} & 4jpx & DYLEA & 4.35 & \underline{\textbf{2.53}} & 4.64 & 4.64 \\
1ppi & PWWER & 2.68 & 3.44 & \underline{\textbf{2.20}} & 3.53 & 4jpy & YLEAY & 3.85 & 2.35 & \underline{\textbf{1.87}} & 4.26 \\
1qin & QQTML & \underline{\textbf{2.75}} & 4.79 & 6.29 & 2.90 & 4mc1 & DTGAD & 3.90 & \underline{\textbf{2.50}} & \underline{\textbf{2.50}} & 2.66 \\
1yc4 & ELISN & 4.09 & \underline{\textbf{3.72}} & \underline{\textbf{3.72}} & \underline{\textbf{3.72}} & 4mo4 & NIGGF & 2.00 & \underline{\textbf{1.91}} & 3.52 & 2.48 \\
1zsf & LLDTG & 2.21 & \underline{\textbf{1.66}} & 5.00 & \underline{\textbf{1.66}} & 4q87 & SLTTP & 3.64 & \underline{\textbf{1.18}} & \underline{\textbf{1.18}} & 1.89 \\
2avo & LIDTG & 3.37 & \underline{\textbf{1.76}} & 3.98 & 3.98 & 4tmk & IEGLE & 4.35 & \underline{\textbf{2.11}} & 3.19 & 3.19 \\
2bfq & AFPAV & 4.46 & \underline{\textbf{1.96}} & 3.42 & 2.31 & 4xaq & GSYSD & \underline{\textbf{3.24}} & 3.42 & 3.40 & 3.42 \\
2bok & EDACQ & 3.33 & 3.50 & 2.47 & \underline{\textbf{1.92}} & 4y79 & ACQGD & \underline{\textbf{2.81}} & 5.96 & 3.28 & 7.08 \\
2qbs & HCSAG & \underline{\textbf{2.43}} & 4.07 & 3.39 & 3.39 & 4zb8 & GGPNG & 3.36 & \underline{\textbf{0.89}} & 1.14 & 2.74 \\
2v25 & ATFTI & \underline{\textbf{1.66}} & 4.03 & 5.11 & 4.03 & 5c28 & CDLCS & \underline{\textbf{3.53}} & 5.17 & 6.14 & 5.17 \\
2vwo & DACQG & 3.39 & \underline{\textbf{2.78}} & 5.91 & 5.91 & 5cqu & RKLGR & 4.83 & \underline{\textbf{2.76}} & 3.66 & 3.66 \\
2xxx & GAVED & 3.75 & \underline{\textbf{3.45}} & 5.94 & \underline{\textbf{3.45}} & 5cxa & FDGKG & 5.22 & \underline{\textbf{2.16}} & \underline{\textbf{2.16}} & 5.88 \\
3b26 & LISNS & 3.63 & 3.61 & 2.34 & \underline{\textbf{1.48}} & 5kqx & LLNTG & 3.77 & \underline{\textbf{2.48}} & 3.16 & 6.96 \\
3ckz & VKDRS & \underline{\textbf{0.83}} & 4.14 & 3.08 & 6.01 & 5kr2 & LNTGA & 4.95 & \underline{\textbf{4.07}} & 4.22 & \underline{\textbf{4.07}} \\
3d7z & YLVTH & 5.48 & \underline{\textbf{2.68}} & 3.79 & 3.79 & 5nkb & MIITE & \underline{\textbf{5.43}} & 6.89 & 7.30 & 7.30 \\
3d83 & LVTHL & 3.75 & 2.43 & \underline{\textbf{1.79}} & 2.24 & 5nkc & IITEY & 4.28 & \underline{\textbf{1.85}} & 1.99 & \underline{\textbf{1.85}} \\
3dx3 & HNDPG & \underline{\textbf{2.30}} & 3.68 & 3.20 & 3.20 & 5nkd & ITEYM & 4.10 & \underline{\textbf{2.24}} & \underline{\textbf{2.24}} & \underline{\textbf{2.24}} \\
3eax & RYRDV & 2.06 & \underline{\textbf{0.43}} & 2.53 & 3.05 & 5tya & LTTPP & 3.67 & 5.51 & \underline{\textbf{1.53}} & 1.62 \\
3ibi & IQFHF & \underline{\textbf{1.91}} & 6.44 & 6.56 & 6.44 & 6czf & LRKAN & \underline{\textbf{2.91}} & 6.97 & 6.53 & 3.20 \\
3nxq & VCHAS & \underline{\textbf{2.97}} & 3.95 & 3.95 & 3.02 & 6ezq & AKQRL & 3.86 & 5.30 & \underline{\textbf{3.10}} & \underline{\textbf{3.10}} \\
3s0b & GIKAV & \underline{\textbf{2.89}} & 4.81 & 5.93 & 4.81 & 6g98 & RNNGH & 3.69 & \underline{\textbf{0.85}} & 5.37 & 2.03 \\
3tcg & IEGVP & \underline{\textbf{1.71}} & 6.03 & 1.91 & 4.74 & 6p86 & VYSSG & \underline{\textbf{4.10}} & 5.80 & 7.55 & 7.55 \\
3vf7 & LDTGA & 3.79 & 3.85 & 3.85 & \underline{\textbf{1.80}} & 6udv & SLSRV & 5.58 & 2.53 & \underline{\textbf{2.38}} & \underline{\textbf{2.38}} \\
4aoi & VVLPY & 3.93 & 1.84 & \underline{\textbf{1.40}} & 1.84 & \multicolumn{2}{>{\cellcolor{gray!20}}l}{\textit{RMSD $<$ QDB}} & \cellcolor{gray!20}-- & \cellcolor{gray!20}32 & \cellcolor{gray!20}28 & \cellcolor{gray!20}28 \\
\bottomrule
\end{tabular}
\end{table}

\subsection{Generalization to Unseen Sequences}
\label{sec:exp2}

The QDockBank results establish that the closed-loop agent improves on the fixed-parameter baseline on roughly half or more of targets per model, with the hardest cases benefiting most. A natural question is whether this advantage extends to sequences outside the benchmark, where no reference structure or ligand-derived signal is available. Critically, the unseen-sequence set uses a minimal-prompt configuration: the designer receives only composition, flexibility, and MJ interaction summaries, with no few-shot examples, penalty ranges, or suggested heuristics. This design isolates whether the LLM can \emph{use feedback across cycles} rather than rely on prompt scaffolding. Because each model chooses a different Hamiltonian, absolute energy should be interpreted cautiously across models; the main question is which model is best at \emph{iterative refinement}.

\textit{Dataset.}
We construct 100 unseen 5-residue peptide sequences with no overlap with QDockBank, selected via a coverage-optimization procedure that maximizes compositional diversity across residue types, dipeptide pairs, and interaction families. The resulting 100 sequences span hydrophobic-rich, polar, charged, and glycine/proline-containing compositions, ensuring that the agent is tested across a broad range of biophysical contexts.

\subsubsection{Reliability gains from the feedback loop}

To quantify the value of the closed-loop mechanism, we compare the agent (best-of-3 cycles) against an LLM-only baseline that runs the same LLM for the same number of independent cycles but without receiving feedback from previous iterations. We report structural validity (defined in Section~\ref{sec:setup}) and \emph{recovery rate}: the fraction of proteins whose cycle-1 structure fails the validity check but whose best-cycle structure passes. Across all 6 configurations (3 models $\times$ 2 noise presets, 600 protein-experiments in total), the agent achieves \textbf{98.7\%} structural validity compared with \textbf{87.5\%} for LLM-only---a gain of \textbf{11.2 percentage points}. All three controllers improve substantially: GPT-5.2 reaches \textbf{100\%} (200/200), GPT-4o \textbf{97.5\%}, and Qwen3.5 \textbf{98.5\%}. Energy range narrows by \textbf{52\%} (0.68 vs.\ 1.41; Wilcoxon $p < 0.001$ in all 6 configurations). Among proteins with invalid cycle-1 structures, the agent recovers \textbf{87\%} (53/61) by best-of-3 cycles and \textbf{95\%} (57/60) by best-of-5.

E/P improves from cycle~1 to cycle~3 in all \textbf{6/6} agent configurations (Wilcoxon $p < 0.05$), while LLM-only shows no directed trend (\textbf{0/6}), replicating the QDockBank behavioral finding on an independent dataset. AlphaFold Server reports high local confidence on these sequences (mean pLDDT $= \textbf{94.3}/100$) but extremely low global fold confidence (pTM $= \textbf{0.03}$--$\textbf{0.04}$, on a 0--1 scale), consistent with locally plausible geometry without a stable global fold---the regime where iterative refinement of local constraints is most relevant.

\begin{figure}[t]
\centering
\includegraphics[width=\columnwidth]{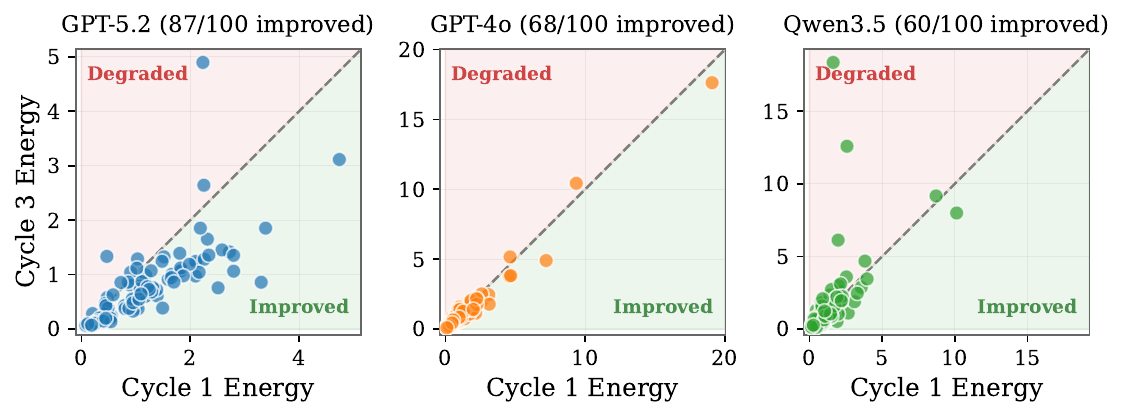}
\caption{Per-sequence energy comparison between cycle~1 and cycle~3 for the 100 unseen sequences (large noise). Points below the diagonal indicate within-model improvement.}
\label{fig:c1_vs_c3}
\end{figure}

\begin{table*}[t]
\caption{Per-cycle VQE statistics on the 100 unseen sequences. $\downarrow$ = lower is better. ``Improved'' is the within-model fraction of proteins with lower cycle-3 than cycle-1 energy; it can exceed 50\% even when the mean increases, because outliers dominate the mean while most individual proteins still improve.}
\label{tab:noise_energy}
\centering
\begin{tabular}{l l rrrc rrrc}
\toprule
& & \multicolumn{4}{c}{Small Noise} & \multicolumn{4}{c}{Large Noise} \\
\cmidrule(lr){3-6} \cmidrule(lr){7-10}
Model & Metric & Cycle 1 & Cycle 2 & Cycle 3 & Improved & Cycle 1 & Cycle 2 & Cycle 3 & Improved \\
\midrule
\multirow{5}{*}{GPT-5.2}
  & Final energy $\downarrow$  & 1.528  & 1.072  & \textbf{0.874}  & \multirow{5}{*}{82\%} & 1.182  & 0.936  & \textbf{0.757}  & \multirow{5}{*}{87\%} \\
  & Energy range $\downarrow$  & 1.137  & 0.746  & \textbf{0.648}  &                       & 0.957  & 0.688  & \textbf{0.568}  & \\
  & Conv.\ rate $\downarrow$   & 0.0193 & 0.0125 & \textbf{0.0109} &                       & 0.0160 & 0.0116 & \textbf{0.0096} & \\
  & Total penalty              & 3.42   & 2.95   & 2.61            &                       & 2.92   & 2.77   & 2.45            & \\
  & Rama favored (\%)          & 98.0   & 97.0   & 96.3            &                       & 98.7   & 98.3   & 97.3            & \\
\midrule
\multirow{5}{*}{GPT-4o}
  & Final energy $\downarrow$  & 1.227  & 1.192  & 1.149  & \multirow{5}{*}{61\%} & 1.471  & 1.331  & 1.321  & \multirow{5}{*}{68\%} \\
  & Energy range $\downarrow$  & 0.997  & 0.977  & 0.859  &                       & 1.086  & 1.041  & 1.008  & \\
  & Conv.\ rate $\downarrow$   & 0.0166 & 0.0166 & 0.0145 &                       & 0.0183 & 0.0175 & 0.0169 & \\
  & Total penalty              & 3.10   & 3.07   & 3.08   &                       & 3.57   & 3.43   & 3.44   & \\
  & Rama favored (\%)          & 96.7   & 97.3   & 96.3   &                       & 95.3   & 96.7   & 96.3   & \\
\midrule
\multirow{5}{*}{Qwen3.5}
  & Final energy $\downarrow$  & 1.199  & 1.293  & 1.273  & \multirow{5}{*}{59\%} & 1.385  & 1.511  & 1.617  & \multirow{5}{*}{60\%} \\
  & Energy range $\downarrow$  & 0.941  & 0.947  & 0.888  &                       & 1.082  & 1.119  & 1.208  & \\
  & Conv.\ rate $\downarrow$   & 0.0159 & 0.0158 & 0.0149 &                       & 0.0181 & 0.0189 & 0.0205 & \\
  & Total penalty              & 3.22   & 3.66   & 3.75   &                       & 3.44   & 4.14   & 4.51   & \\
  & Rama favored (\%)          & 96.7   & 96.3   & 97.7   &                       & 96.0   & 98.0   & 97.3   & \\
\bottomrule
\end{tabular}
\end{table*}

\subsubsection{Controller profiles and penalty mechanism}

The clearest pattern on the unseen-sequence set is that \textbf{GPT-5.2 is the strongest refinement model} (Table~\ref{tab:noise_energy}). Across both noise settings, GPT-5.2 improves monotonically across cycles. Under large noise, mean final energy decreases from 1.182 in cycle~1 to 0.757 in cycle~3, and \textbf{87\%} of sequences improve; under small noise, it decreases from 1.528 to 0.874, with \textbf{82\%} improving. GPT-4o shows weaker refinement (68\% and 61\% improving), while Qwen3.5 is the least consistent in energy trajectory (\textbf{59--60\%} improving) but uniquely improves structural validity---a different optimization strategy that prioritizes constraint satisfaction over energy minimization. Because each model chooses a different Hamiltonian, absolute energy comparison across models should be interpreted cautiously; the within-model improvement rate is the primary signal. This hierarchy suggests that while the closed-loop framework provides the foundation, the quality of the LLM controller determines how much of that potential is realized.

Across all three controllers, later cycles consistently produce tighter energy ranges and lower convergence rates than cycle~1, indicating that the feedback loop yields smoother optimization landscapes regardless of the backend. The per-protein distribution (Figure~\ref{fig:c1_vs_c3}) reveals that the aggregate numbers in Table~\ref{tab:noise_energy} hide distinct controller behaviors: GPT-5.2's improvements are broadly distributed across the dataset rather than driven by a few large gains; GPT-4o's points cluster tightly near the diagonal, reflecting small, consistent adjustments; and Qwen3.5 shows large outliers in both directions, with some proteins exhibiting substantial energy degradation.

\begin{figure}[t]
\centering
\includegraphics[width=\columnwidth]{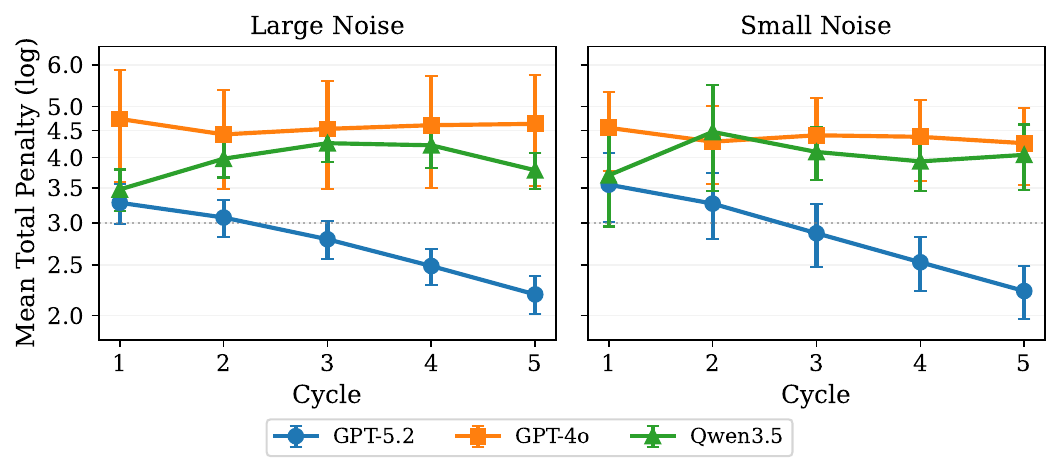}
\caption{Mean total penalty ($\lambda_{\text{chiral}} + \lambda_{\text{back}} + \lambda_1$) across 5 cycles (log scale). GPT-5.2 consistently reduces penalties, GPT-4o plateaus, and Qwen3.5 tends to increase.}
\label{fig:penalty_trajectory}
\end{figure}

The refinement advantage of GPT-5.2 is reflected in its penalty updates (Figure~\ref{fig:penalty_trajectory}). Under large noise, \texttt{penalty\_back} decreases in \textbf{84\%} of sequences from cycle~1 to 2 and \textbf{81\%} from cycle~2 to 3. Crucially, this reduction does not come at the cost of structural quality: Ramachandran favored percentages remain above 96\% across cycles (Table~\ref{tab:noise_energy}), indicating that the backbone constraint is already being satisfied when the penalty is relaxed. Qwen3.5, in contrast, increases total penalty on average across cycles. The controller hierarchy in Table~\ref{tab:noise_energy} aligns with these penalty-update patterns.

\subsubsection{Sustained refinement through five cycles}

\begin{figure}[t]
\centering
\includegraphics[width=\columnwidth]{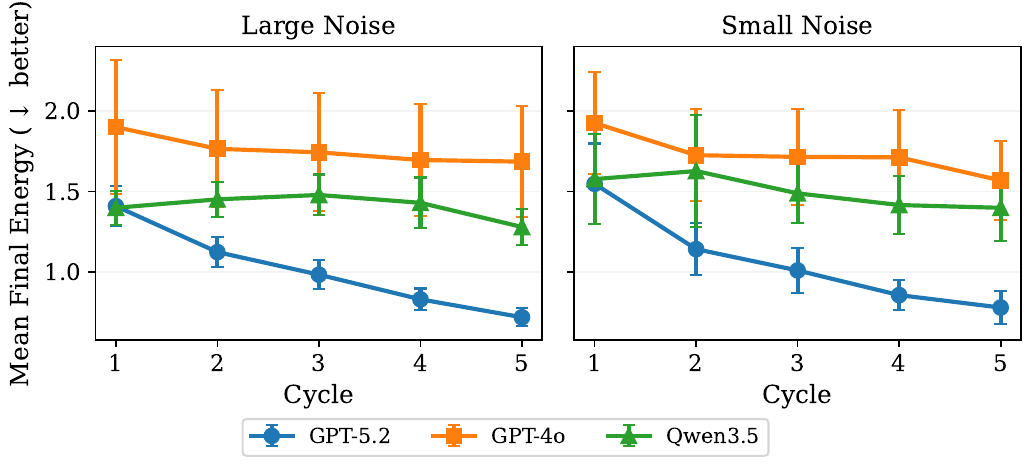}
\caption{Energy trajectory extended to five optimization cycles. GPT-5.2 continues to improve through cycles~4--5, demonstrating sustained closed-loop value when feedback interpretation is consistent.}
\label{fig:energy_5c}
\end{figure}

To verify that the directed optimization observed at 3 cycles reflects a sustained behavioral pattern rather than an early-cycle artifact, we extended all models to five cycles on the same 100 unseen sequences. Figure~\ref{fig:energy_5c} shows that GPT-5.2 maintains a consistent downward energy trajectory through cycles~4 and~5, while others plateau. The best structure does not always come from the final cycle---the feedback loop enables exploration across cycles, and best-of-$k$ selection captures the strongest result regardless of which cycle produced it. Structural validity exhibits a U-shaped trajectory: validity dips in mid-cycle as the agent explores lower penalties, then recovers by cycle~5 (best-of-5 pooled: \textbf{99.5\%}, GPT-5.2: \textbf{100\%}). This pattern indicates self-correcting behavior---the agent first explores aggressively, then learns the structural validity boundary and adjusts accordingly. The controller ranking is preserved across both noise presets with no rank inversions.

\subsection{Ablation Study}
\label{sec:ablation}

We decompose the sources of improvement using three ablation dimensions.

\subsubsection{Closed-Loop Feedback vs.\ Single Cycle}

To isolate the value of iterative feedback, we compare cycle-1 energy (no feedback) against the best energy across all 3 cycles (with feedback) for each protein. All nine configurations (3 models $\times$ 3 noise levels including noiseless) show statistically significant improvement (Wilcoxon $p < 10^{-11}$). The magnitude varies by model: GPT-5.2 achieves \textbf{39--43\%} energy reduction, GPT-4o \textbf{11--17\%}, and Qwen3.5 \textbf{14--24\%} (Table~\ref{tab:ablation_feedback}). Even Qwen3.5, which shows net energy \emph{increase} at cycle~3 on average (Table~\ref{tab:noise_energy}), still achieves a lower \emph{best} energy across cycles than its cycle-1 starting point. The feedback loop enables exploration even when the average trajectory of the model is suboptimal.

\begin{table}[t]
\caption{Single cycle vs.\ closed-loop ablation. C1 Mean = mean VQE final energy at cycle~1; Best Mean = mean of the lowest energy across 3 cycles per protein; $\Delta$\% = relative reduction. All 9 model--noise configurations are significant (Wilcoxon $p < 10^{-11}$).}
\label{tab:ablation_feedback}
\centering
\small
\begin{tabular}{l l r r r}
\toprule
Model & Noise & C1 Mean & Best Mean & $\Delta$\% \\
\midrule
GPT-5.2  & large & 1.548 & 0.883 & $-$43\% \\
GPT-5.2  & small & 1.175 & 0.721 & $-$39\% \\
GPT-5.2  & none  & 3.419 & 2.190 & $-$36\% \\
\midrule
GPT-4o   & large & 1.943 & 1.605 & $-$17\% \\
GPT-4o   & small & 1.369 & 1.156 & $-$16\% \\
GPT-4o   & none  & 0.470 & 0.419 & $-$11\% \\
\midrule
Qwen3.5 & large & 1.233 & 0.949 & $-$23\% \\
Qwen3.5 & small & 1.199 & 0.913 & $-$24\% \\
Qwen3.5 & none  & 1.876 & 1.605 & $-$14\% \\
\bottomrule
\end{tabular}
\vspace{-5pt}
\end{table}

\subsubsection{Effect of Noise Level}

Holding the model constant and varying noise, GPT-5.2 shows a consistent improvement pattern: 82\% of proteins improve under no noise, 87\% under small noise, and 89\% under large noise. Counter-intuitively, noisier conditions produce \emph{higher} improvement rates, likely because noise roughens the energy landscape, creating more room for penalty refinement to steer the optimizer away from poor local minima. Qwen3.5 shows the opposite pattern: only 37\% improve under no noise, rising to 64\% under large noise, where noise-induced energy fluctuations create a more forgiving optimization landscape. GPT-4o falls between, with 45--67\% improvement rates. This suggests that the agent's feedback strategy interacts with the noise profile: a strong model becomes \emph{more} effective under noise, while a weaker model is partially rescued by it.

\subsubsection{Penalty Dynamics}
\label{sec:ablation-penalty}

Penalty magnitude is the dominant optimization lever. Across all three models under large noise, \texttt{penalty\_back} is strongly associated with VQE final energy ($R^2 > 0.99$), although part of this relationship is definitional because \texttt{penalty\_back} enters the Hamiltonian directly as the coefficient of $H_{\text{back}}$ (Eq.~1). The more informative test is whether \emph{changes} in penalty track \emph{changes} in energy across cycles, which cannot be explained by the Hamiltonian definition alone. The change in total penalty from cycle~1 to cycle~3 explains $R^2 = 0.72$ of the energy improvement for GPT-5.2, but only $R^2 = 0.41$ for Qwen3.5, because the penalty changes of Qwen3.5 are not consistently directional. Given this dominance, the agent's contribution reduces to a directional question: does it decrease penalties when structural constraints are already satisfied? Penalty direction predicts outcome: for every model and noise level, proteins where \texttt{penalty\_back} decreases show energy improvement (mean $\Delta E = +0.34$ to $+1.39$), while proteins where it increases show degradation ($\Delta E = -0.17$ to $-5.06$). GPT-5.2 decreases \texttt{penalty\_back} in \textbf{78--89\%} of cases, yielding \textbf{82--89\%} improvement; GPT-4o decreases in 32--65\%, yielding 45--67\% improvement; Qwen3.5 decreases in 36--59\%, yielding 37--64\% improvement. This confirms the premise established in Section~\ref{sec:background}: penalty weight choice is the dominant lever in quantum protein optimization. Beyond magnitude, the agent also produces more balanced penalty distributions: the standard deviation across the three penalty weights is significantly lower than LLM-only in 10 of 12 experiments, indicating that the feedback loop learns to distribute constraint weights more uniformly.

\subsection{Discussion}
\label{sec:discussion}

The experiments demonstrate clear process-level improvements, including more directed optimization, higher structural validity, and smoother energy landscapes, even though endpoint-level RMSD gains are not yet distinguishable at the 5-residue scale. The ablation shows that penalty magnitude is the dominant control variable, and that the feedback loop is what makes this variable effective. Because the refinement loop is modular and independent of the quantum backend, the workflow can scale naturally with available hardware. For 6 to 8 residue fragments, tensor-network simulation via Qiskit Aer’s matrix product state backend can replace statevector simulation, while for 9+ residue fragments the same pipeline can run on IBM Quantum hardware backends~\cite{chow2021ibm,kim2023evidence} by swapping only the VQE execution layer. The agent loop, metric computation, and structural validation remain unchanged. From this perspective, the current 5-residue setting is best understood as a controlled proof of concept rather than a structural limitation of the framework. An important next step is to extend the agent beyond penalty refinement so that it can also design sequence-specific quantum circuits and execution settings, such as ans\"atz choice, circuit depth, entanglement structure, shot allocation, and optimizer configuration. Such a direction would enable a more adaptive workflow and research opportunities in which both the optimization objective and the quantum strategy are tailored to each unique sequence.

\section{Conclusion}

We presented QFoldAgent, a closed-loop multi-agent framework that adaptively refines Hamiltonian penalties for quantum-classical protein structure prediction. Through a Designer Agent and a Feedback Agent, the framework first proposes sequence-conditioned penalty settings and then refines them across cycles using structural validation and optimization feedback. On 55 QDockBank fragments, the framework improves RMSD over the fixed-parameter baseline for each model individually, with the largest gains on the hardest targets. Behavioral analysis confirms that these gains arise from directed refinement: E/P improves in 5/6 agent configurations while LLM-only controls show no trend (0/6). On 100 unseen sequences, the advantage generalizes: structural validity rises from 87.5\% to 98.7\%, with 87\% recovery of initially invalid structures and sustained refinement behavior confirmed through five cycles.

We release the complete evaluation suite---55 ground-truth targets, 100 coverage-optimized unseen sequences, evaluation scripts, and baseline results from three controllers under two noise regimes---as a reusable protocol for comparing future LLM controllers in quantum optimization workflows.

\bibliographystyle{IEEEtran}
\bibliography{references}

\end{document}